\documentclass[10pt,twocolumn,letterpaper]{article}

\usepackage{cvpr}
\usepackage{times}
\usepackage{epsfig}
\usepackage{graphicx}
\usepackage{subcaption}
\usepackage{amsmath}
\usepackage{amssymb}
\usepackage{booktabs}
\usepackage{multicol}

\usepackage[pagebackref=true,breaklinks=true,letterpaper=true,colorlinks,bookmarks=false]{hyperref}

\cvprfinalcopy 

\def\cvprPaperID{****} 

\ifcvprfinal\fi
\begin{document}

\title{Benchmarking Graph Neural Networks on Link Prediction}





\author{
Xing Wang 


\qquad  \qquad \quad
Alexander Vinel \\


{\tt\small xzw0005@auburn.edu \qquad azv0019@auburn.edu} \\

Auburn University, AL, 36849, USA



}

\maketitle

\begin{abstract}
In this paper, we benchmark several existing graph neural network (GNN) models on different datasets for link predictions. In particular, the graph convolutional network (GCN), GraphSAGE, graph attention network (GAT) as well as variational graph auto-encoder (VGAE) are implemented dedicated to link prediction tasks, in-depth analysis are performed, and results from several different papers are replicated, also a more fair and systematic comparison are provided.
Our experiments show these GNN architectures perform similarly on various benchmarks for link prediction tasks.

\end{abstract}


\section{Introduction}

Our primary focus in this paper is link prediction, which aims to infer unobserved/missing links or predicting the future ones based on the connections of currently observed partial networks. Link prediction is a fundamental problem in network science with tremendous real-world applications \cite{lu2011link}. 


Over the years, there are many link prediction methods have been developed. Local link prediction methods based on the assumption that two nodes are more likely to be connected if they have many common neighbors \cite{zhou2009predicting}, but generally obtained low accuracy especially on large and sparse graphs. The global structural similarity was also taken consideration \cite{liu2013hidden}, but is generally computationally infeasible for large graphs. There are also probabilistic approaches assuming a known prior structure of the graph, such as hierarchical or circles structures \cite{clauset2008hierarchical}, though the accuracy was not satisfactory. And there has been a surge of algorithms that make link prediction through representation learning that learns low dimensional embeddings for nodes, most of these methods are based on skip-gram or matrix factorization, examples include DeepWalk \cite{perozzi2014deepwalk}, node2vec \cite{grover2016node2vec}, etc. These algorithms are basically unsupervised learning with no supervised information during training, although the learned representations can assist downstream tasks such as node classification, graph classification, as well as link prediction to achieve higher accuracy.

As the graph neural networks (GNNs) have been an emerging research area in recent years, significant advances and various architectures are proposed and developed. Lots of the new models are evaluated on the citation network datasets (Cora, CiteSeer, PubMed). However, different training/test splits were often deployed across studies \cite{mernyei2020wiki}, which is challenging to make fair comparisons among different methods. 
Also, all three citation networks are derived from the same domain, with similar structural properties.
Furthermore, when most of the models developed, they were primarily tested on tasks such as node classification and graph classification, not all performed link prediction. For instance, in \cite{kipf2016semi}, \cite{hamilton2017inductive} and \cite{velivckovic2017graph} that we will illustrate in Section \ref{sec:approach}, only node classification and graph classification tasks were benchmarked, but no link prediction task was evaluated. \cite{kipf2016variational} experimented link prediction in the citation networks, and compared only with baselines that are not recently developed, such as spectral clustering \cite{tang2011leveraging} and DeepWalk \cite{perozzi2014deepwalk}. 
\cite{mernyei2020wiki} introduced an additional dataset for various GNN benchmarks, but for link prediction task, only variational graph auto-encoder (VGAE) was evaluated and compared with the multi-layer perceptron (MLP) baseline. 
In this paper, we aim at providing a fair comparison on several most influential GNN architectures that were developed in recent years, evaluating them on the same datasets, also with consistent training splits of data among different models. Moreover, we will benchmark the models on additional dataset other than the commonly used citation networks. 

The rest of this paper is organized as follows. Section \ref{sec:data} introduces the four datasets we test on.  Section \ref{sec:approach} illustrates several recently developed GNN architectures that we apply and evaluate.  In Section \ref{sec:implement}, implementation details are discussed.  We benchmark the models and show the evaluation results in Section \ref{sec:res}. Finally, we conclude our findings and discuss future work directions in Section \ref{sec:conc}.




\section{Datasets} \label{sec:data}
For this paper, we used the following datasets:

\begin{itemize}
    \item \textbf{Cora}: This dataset is a citation network that is used as a standard benchmark dataset. The network has 2708 nodes (scientific publications) which are classified into one of seven classes. The citation network contains 5429 links (undirected edges) between nodes. Each node is described by a 0/1 bag-of-words vector with a dictionary size of 1433 \cite{linqsdata}.

    \item \textbf{CiteSeer}: This dataset is another citation network similar to Cora. It contains 3327 nodes (scientific publications), 4732 edges (links), and 6 classes. Each node in this dataset is also described by a 0/1 bag-of-words vector with a dictionary size of 3703 \cite{linqsdata}.
       
    \item \textbf{PubMed}: This dataset is a citation network of scientific publications pertaining to diabetes from the Pubmed database. It contains 19717 nodes, 44338 edges, and 3 classes. Each node is described by a TF/IDF weighted word vector with a dictionary size of 500 \cite{linqsdata}.
    
    \item \textbf{WikiCS}: This dataset is based on Wikipedia article references. It consists of 11701 nodes (Wikipedia articles on Computer Science), 216213 edges (hyperlinks between articles), and 10 classes. Each node is described by a 300-dimensional vector which was obtained as the average of pretrained GloVe embeddings \cite{mernyei2020wiki}.
\end{itemize}

Cora, CiteSeer, and PubMed are commonly used benchmark datasets for tasks related to graph neural networks such as link prediction, node classification, and graph classification. 
While these three datasets are all citation networks, the Wiki-CS is a newly published dataset in a different domain. It also has different structural properties, the primary one being that there are significantly more edges per node. It has a mean degree of 36.94 while PubMed has a mean degree of 4.5 in spite of having nearly 70\% more nodes \cite{mernyei2020wiki}, thus has a significantly higher connectivity rate and provides a different distribution as a testbed. In this paper we use the abovementioned four dataset to benchmark several recently developed GNN architectures on the link prediction task. 


\section{Approach} \label{sec:approach}

Our primary goal is to benchmark several recently developed GNN models for link prediction task. 
With consistent training splits of data among different methods, our work provides a more fair comparison.
In particular, the following four models will be implemented and evaluated for link prediction tasks on the datasets we discussed earlier in this paper. 

\subsection{Graph Convolutional Networks (GCN)} \label{sec:gcn}
GCN \cite{kipf2016semi} borrows the concept of convolution from the convolutional neural network (CNN) and convolve the graph directly according to the connectivity structure of the graph as the filter to perform neighborhood mixing. The architecture can be summarized concisely as
$$ H^{l+1} = \sigma (\tilde{D}^{-1/2} \tilde{A}^{-1/2} \tilde{D}^{-1/2} H^{(l)} W^{(l)}), $$
where $A$ and $D$ are the adjacency matrix and the degree matrix of the graph, respectively. 
Two tricks were applied that help the GCN being successful: a self-connection is added to each node in the adjacency matrix, and then the adjacency matrix is normalized according to the degrees, so that we obtained $\tilde{A}$ and $\tilde{D}$. $H^{(l)}$ contains the embedding of the nodes in $l^{th}$ layer, $W^{(l)}$ denotes the weight matrix for this layer, and $\sigma$ is the nonlinearity. 

\subsection{GraphSAGE} \label{sec:sage}
GraphSAGE \cite{hamilton2017inductive} is an inductive learning algorithm for GNNs, meaning that instead of applying the whole adjacency matrix information among all node, it learns aggregator functions that can induce the embedding of a new node given its features and neighborhood information without retraining of the entire model. To learn the embeddings with aggregators, a neighborhood embedding  would be mapped with the aggregation function and concatenated with existing embedding of the node. The concatecated vector would be passed through a GNN layer to update the node embedding, and finally the embeddings are normalized to unit norm. Note the aggregator can be of various form, \cite{hamilton2017inductive} proposed the options of the mean (as used in GCN), pooling, as well as LSTM. The sampling from neighbors strategy in GraphSAGE yields impressive performance on node labeling task over several large-scale networks. 

To learn the weights of aggregators and embeddings, the loss function of GraphSAGE is defined as
$$ \mathcal{L} = -\log \sigma (z_u^T z_v) - Q \mathbb{E}_{v_n\sim P_n(v)} \log \sigma (-z_u^T z_{v_n}), $$
where $\sigma$ represents the sigmoid function, $u$ and $v$ are two neighbors, while $v_n$ is a negative sample, and $Q$ is the number of negative samples. The first term aims at maximizing the similarity between embeddings of $u$ and $v$, while the second term tries to set apart embeddings of negative samples.  

\subsection{Graph Attention Networks (GAT)}
In recent years, attention mechanisms have become state-of-the-art (SOTA) for many tasks on sequential data, such as in natural language processing. GAT \cite{velivckovic2017graph} combined attention mechanisms with GNN, aiming at more effective learning power to the neighborhoods' features. As the building block of GAT, the graph attention layer plays the role of aggregation function for the GNN. 
It first applies a shared linear transformation to every node, parameterized by the weight matrix $\mathbf{W}\in \mathbb{R}^{F'\times F}$. Then self-attention is performed on the nodes where a shared attention mechanism is used to calculate the attention coefficients that capture the importance of node $j$'s features to node $i$, i.e., 
$$e_{ij} = a(\mathbf{W}h_i, \mathbf{W}h_j).$$
GAT normalized the coefficients across $j$ using a softmax function. As a result, the attention mechanism is a single-layer network parameterized by a weight vector $a\in \mathbb{R}^{2F'}$, followed by a nonlinear activation (e.g., LeakyReLU), and obtains the normalized coefficients as
$$ \alpha_{ij} = \dfrac{\exp \Big( \text{LeakyReLU} (a[\mathbf{W}h_i, \mathbf{W}h_j]) \Big)}{\sum_{k\in N_i} \exp \Big( \text{LeakyReLU} (a[\mathbf{W}h_i, \mathbf{W}h_k]) \Big) }. $$

These attention coefficients are used to compute a linear combination of the neighbors' features to obtain aggregated features for each node, i.e., 
$$h'_i = \sum_{k\in N_i} \alpha_{ij} \mathbf{W} h_k.$$

And finally, multi-head attention (i.e., a set of independent attention mechanism) is often employed to stabilize the learning process of self-attention. 

\subsection{Variational Graph Auto-Encoders}
Kipf and Welling introduced the Variational Graph Auto-Encoders (VGAE) framework for unsupervised learning on graph-structured data \cite{kipf2016variational}. This framework is an elegant extension of the Variational Auto-Encoders framework \cite{kingma2014autoencoding} to apply to graph-structured data. 
The VGAE uses a graph convolutional network \cite{kipf2016semi} as the encoder to map the input graph's node features ($X$) into the latent representation, followed by an inner product decoder to generate the conditional probabilities of the adjacency matrix ($A$) given the latent representation.
The hidden layer of the encoder can be a GCN layer as we described in \ref{sec:gcn}:
\begin{gather*}
\bar{X} = GCN(X, A) = ReLU(\bar{A} X W_0) \\
\tilde{A} = D^{-\frac{1}{2}} A D^{-\frac{1}{2}}
\end{gather*}
where $\tilde{A}$ is the symmetrically normalized adjacency matrix \cite{Fanghao}.
By applying reparametrization trick \cite{kingma2014autoencoding}, the latent distribution produced by the encoder can be written as
\begin{align*}
\mu &= GCN_{\mu}(X, A) = \tilde{A} \bar{X} W_1 \\
\log\sigma &= GCN_{\sigma}(X, A) = \tilde{A} \bar{X} W_1 \\
Z &= \mu + \sigma * \epsilon
\end{align*}
where $ \epsilon \sim  \mathcal{N}(0, 1) $.

Thus the encoder is given by \cite{kipf2016variational} as
$$ q(z_i | X, A)=\mathcal{N}(z_i | \mu_i, diag(\sigma_i^2) ) $$
and the decoder is 
$$ p(A_{ij}=1 | z_i, z_j) = \sigma(z_i^T z_i) $$

Finally, the loss function 
is a combination of the reconstruction loss and the KL-divergence:
$$ \mathcal{L} = \mathbb{E}_{q(Z|X,A)} [ \log{p(A|Z)} ] - KL[ q(Z|X,A) \| p(Z) ] $$

\section{Implementation Details} \label{sec:implement}
The four GNN models illustrated in Section \ref{sec:approach} are benchmarked for link prediction tasks on the four dataset mentioned in Section \ref{sec:data}. In this section, we describe the architectures for the four models. 

We constructed GCN, GraphSAGE, GAT and VGAE models for link prediction also with PyTorchGeometric \cite{Fey/Lenssen/2019}. We used the same default architectures as the original implementations of corresponding papers.
\footnote{
The PyTorchGeometric implementation of GCN is adapted from \url{https://github.com/rusty1s/pytorch_geometric/blob/master/examples/link_pred.py}, and the implementation of other models follows the paradigm with the change of network architectures.
}
For GCN, we built a two level graph convolutional layers, with 128 and 64 filters, respectively. The two graph convolutional layers play a similar role as the encoder in VGAE. The outputs of the graph convolutional layers are used for predict the label of a link between two nodes similar to the decoder, as we will describe later. As for GraphSAGE, we first implemented the SAGE module as described in Section \ref{sec:sage}, and we build the encoder-like module for our GraphSAGE link prediction model as two such SAGE layers, each with 64 hidden channels. 
We built two graph attention layers for GAT, with a ReLU nonlineararity in between. The first layer is of multi-headed attention using 8 heads, each consists of 8 filters, and with a dropout rate of 0.6, while the second graph attention layer has a single attention head consists of 16 filters. 
And finally, for the VGAE, we defined the encoder as a hidden graph convolutional layer consists of 32 convolutional filters, followed by two graph convolutional layers representing the mean and log standard deviation of the latent normal distribution respectively (as per \cite{kipf2016variational}). The latent dimension was set as 16. And the loss function optimized was set to be the sum of the reconstruction loss and KL-divergence loss.

Our implementation of the GCN, GraphSAGE, and GAT models use the same decoder-like module to output the predicted probabilities of the existence for a link. Similar to the decoder in VGAE, we use dot product to calculate the similarities of linkage for corresponding nodes pairs. During training, negative sampling strategy was deployed to generate negative samples to match the positive samples (i.e., existing edges in the original graphs) within each minibatch, and binary cross entropy loss was backpropagated for such a binary classification problem, which is different from VGAE which is a generative model.

\section{Experiments and Results} \label{sec:res}

\begin{table*}
\begin{center}
\begin{tabular}{|c|c|c|c|c|}
\toprule
\textbf{Method} & \textbf{Cora} & \textbf{CiteSeer} & \textbf{PubMed} & \textbf{Wiki-CS} \\
\midrule
GCN
& $0.9273 \pm 0.0051$ 
& $0.8911 \pm 0.0058$ 
& $0.9619 \pm 0.0011$ 
& $0.9682 \pm 0.0030$ 
\\
GraphSAGE
& $0.9091 \pm 0.0038$ 
& $0.8716 \pm 0.0063$ 
& $0.9018 \pm 0.0032$ 
& $0.8758 \pm 0.0423$ 
\\
GAT
& $0.9027 \pm 0.0023$ 
& $0.9084 \pm 0.0035$ 
& $0.9179 \pm 0.0002$ 
& $0.9293 \pm 0.0014$
\\
VGAE
& 0.9179 $\pm$ 0.0013 
& 0.9091 $\pm$ 0.0021
& 0.9293 $\pm$ 0.0028 
& 0.9578 $\pm$ 0.0008 
\\
\bottomrule
\end{tabular}
\end{center}
\caption{AUC comparison for GNN models on four datasets.}
\label{tab:auc}
\end{table*}

\begin{table*}
\begin{center}
\begin{tabular}{|c|c|c|c|c|}
\toprule
\textbf{Method} & \textbf{Cora} & \textbf{CiteSeer} & \textbf{PubMed} & \textbf{Wiki-CS} \\
\midrule
GCN
& $0.9385\pm 0.0043$ 
& $0.9028\pm 0.0050$ 
& $0.9636 \pm 0.0010$ 
& $0.9720 \pm 0.0025$ 
\\
GraphSAGE
& $0.9183 \pm 0.0035$ 
& $0.8863 \pm 0.0052$ 
& $0.9057 \pm 0.0025$ 
& $0.8778 \pm 0.0425$ 
\\
GAT
& $0.9027\pm 0.0023$
& $0.9084\pm 0.0035$ 
& $0.9037 \pm 0.0021$ 
& $0.9277 \pm 0.0013$ 
\\
VGAE
& 0.9297 $\pm$ 0.0013
& 0.9234 $\pm$ 0.0017
& 0.9336 $\pm$ 0.0027
& 0.9623 $\pm$ 0.0008
\\
\bottomrule
\end{tabular}
\end{center}
\caption{Average precision scores comparison for GNN models on four datasets.}
\label{tab:ap}
\end{table*}

We conducted experiments of link prediction with the GCN, GraphSage, GAT, as well as VGAE models on the datasets we described in Section \ref{sec:data} (Cora, CiteSeer, PubMed, Wiki-CS). 
All four datasets are available via Pytorch Geometric  APIs \cite{Fey/Lenssen/2019}.
For each of the datasets, the edges in the graph were split into training/validation/testing sets with a ratio of 85\%-5\%-10\%. As we mentioned above, negative edge samples were used for calculating the loss, by randomly sampling node pairs without an linkage edge with the same number as the positive edges within each mini-batch. All of our models were trained using Adam optimizers \cite{kingma2014adam} for stochastic gradient descent, while the initial learning rates were set to 0.01, with an exception of the GAT task on WikiCS, where the initial learning rate was set to 0.001. All the tasks were trained for 200 epochs, and in order to replicate \cite{mernyei2020wiki}, we run each task for 50 runs so that the results are reported as the average of 50 runs along with 95\% confidence intervals calculated using bootstapping.

We report the area under the ROC curve (AUC) and average precision scores for each of link prediction tasks with the four GNN models. The AUC results are showed in Table \ref{tab:auc}, while Table \ref{tab:ap} reports the average precision scores. From the two tables, we observe that the GCN model performs best on Cora, PubMed, as well as the relatively larger Wiki-CS tasks, while VGAE achieves the best performance on CiteSeer. Overall, the four models perform quite similar on most of the tasks, with the AUC around 90\%. The results obtained by GraphSage may seem a little poorer for CiteSeer and Wiki-CS, not only in terms of smaller mean AUC and average precision, also with larger standard deviation on the metrics, indicating the model is less robust in these tasks. Considering GraphSage models sample from neighbors for each node during propagation over the graph, less robustness could be expected especially for the graph with imbalanced distributions. The GCN and VGAE perform significantly better on Wiki-CS, as the mean AUCs and the average precisions exceed 95\%.

Furthermore, we include the visualizations of the embeddings obtained from each model for the four datasets. The embeddings were obtained as the the latent space encodings for the VGAE model, and as the output of the graph convolutional or attention layers which serves as feature maps for the downstream classifiers in the GCN, GraphSage and GAT models. 
T-SNE algorithm \cite{maaten2008visualizing} was applied to project the high dimensional embeddings into a 2D plot. Figures \ref{fig:cora_vis},  \ref{fig:citeseer_vis},  \ref{fig:pubmed_vis}, and  \ref{fig:wikics_vis} in Appendix show the t-SNE visualization for the embeddings of nodes in Cora, CiteSeer, PubMed, and Wiki-CS datasets, respectively.
We observe that after training on the link prediction tasks, our GNN models are able to generally project the nodes with the same class label into closed neighboring areas of the latent space with fairly reasonable separation between the classes. This demonstrates that the models are able to adequately learn good latent representations to distinguish between the types of labels, particularly for VGAE in spite of that the model being unsupervised. We also observe that for those models perform well on a particular task, the embedding visualization presents a clearer distinction among different classes, such as in Figures \ref{fig:cora_vis}(a), \ref{fig:citeseer_vis}(b), and \ref{fig:wikics_vis}(a). Opposite examples include that was shown in Figure \ref{fig:wikics_vis}(b), where the node embeddings present some spiral patterns, indicating some correlation between the two projected dimensions, which might be a further implication that the GraphSage model is overfitting on this task as the prediction performance is worse that that of other models, and we probably could use this information to monitor the training, tune the hyperparameters, or apply additional regularization such as early stopping.




\section{Conclusions and Discussion} \label{sec:conc}
In this paper, we implemented several existing GNN models, and benchmarked on different datasets for link predictions. We not only 
reproduced the results in \cite{kipf2016variational} and \cite{mernyei2020wiki},
also provided a more fair and systematic comparison.
Our experiments show these GNN architectures perform similarly on other benchmarks for link prediction tasks.
There are several interesting directions to pursue for the future of this paper. First off, the datasets we benchmarked on are still relatively small, in the future we could evaluate the models on much larger graphs, especially with applications in real world. The second interesting direction in which to take this paper would be to implement more recently developed GNN models. Furthermore, we could try to design and develop our own GNN architecture and benchmark on link prediction tasks.

{\small
\bibliographystyle{ieee_fullname}
\bibliography{egbib}
}

\newpage
\appendix
\section{T-SNE Embedding Visualization}


\begin{figure}[ht!]
    \centering
    \begin{subfigure}[b]{0.325\textwidth}
        \centering
        \includegraphics[width=\textwidth]{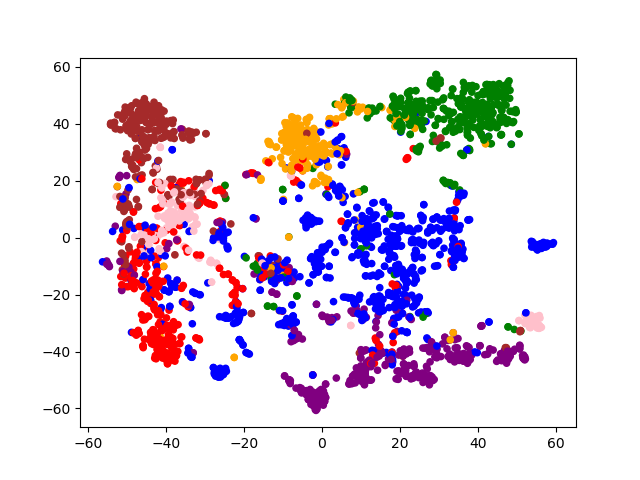}
    \caption{GCN}
    \end{subfigure}
    ~
    \begin{subfigure}[b]{0.325\textwidth}
        \centering
        \includegraphics[width=\textwidth]{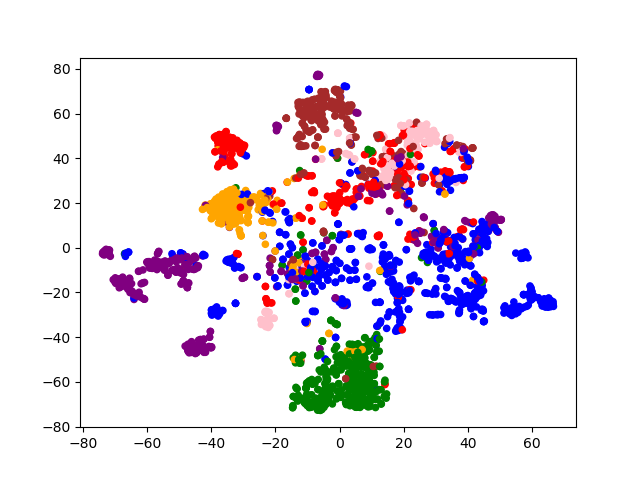}
        \caption{GraphSage}
    \end{subfigure}
    ~
    \begin{subfigure}[b]{0.325\textwidth}
        \centering
        \includegraphics[width=\textwidth]{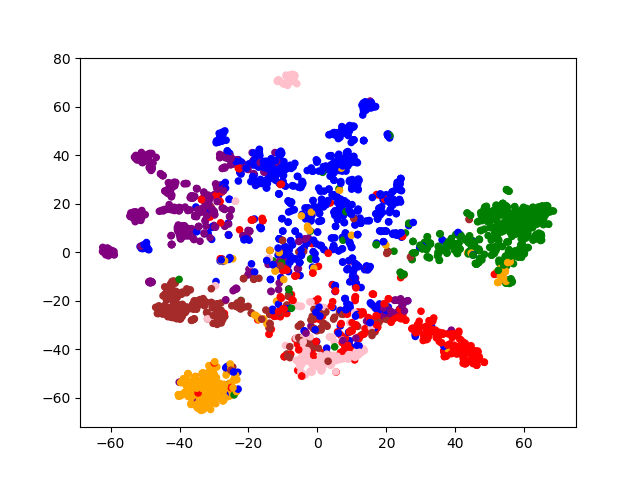}
        \caption{GAT}
    \end{subfigure}
    ~
    \begin{subfigure}[b]{0.325\textwidth}
        \centering
        \includegraphics[width=\textwidth]{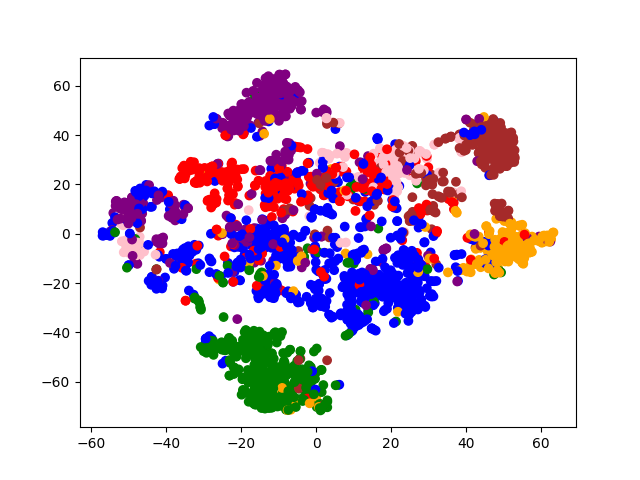}
        \caption{VGAE}
    \end{subfigure}
    \caption{Cora Embeddings Visualization}
    \label{fig:cora_vis}
\end{figure}

\begin{figure}[ht!]
    \centering
    \begin{subfigure}[b]{0.325\textwidth}
        \centering
        \includegraphics[width=\textwidth]{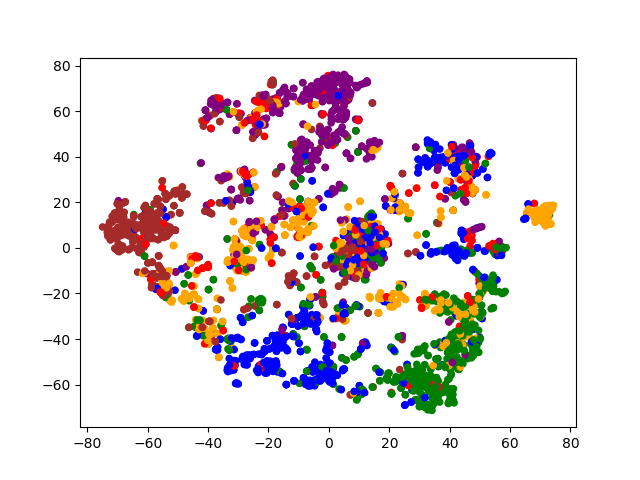}
    \caption{GCN}
    \end{subfigure}
    ~
    \begin{subfigure}[b]{0.325\textwidth}
        \centering
        \includegraphics[width=\textwidth]{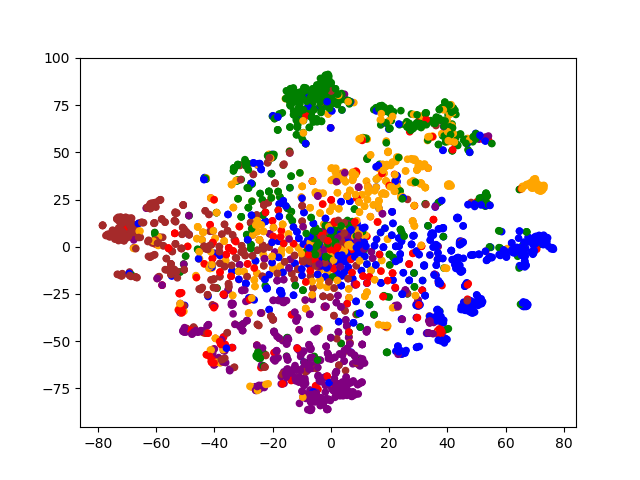}
        \caption{GraphSage}
    \end{subfigure}
    ~
    \begin{subfigure}[b]{0.325\textwidth}
        \centering
        \includegraphics[width=\textwidth]{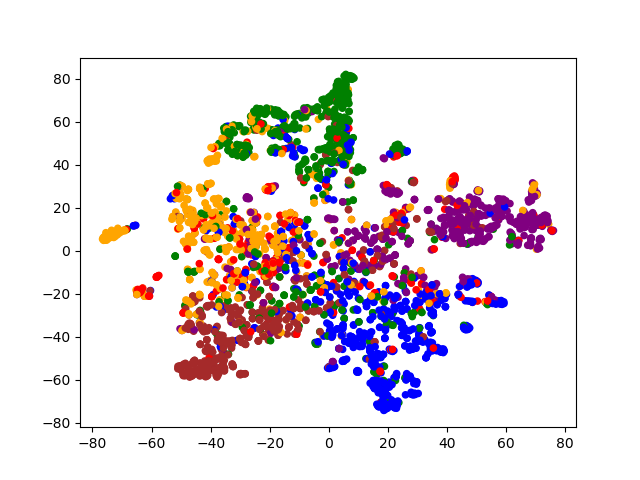}
        \caption{GAT}
    \end{subfigure}
    ~
    \begin{subfigure}[b]{0.325\textwidth}
        \centering
        \includegraphics[width=\textwidth]{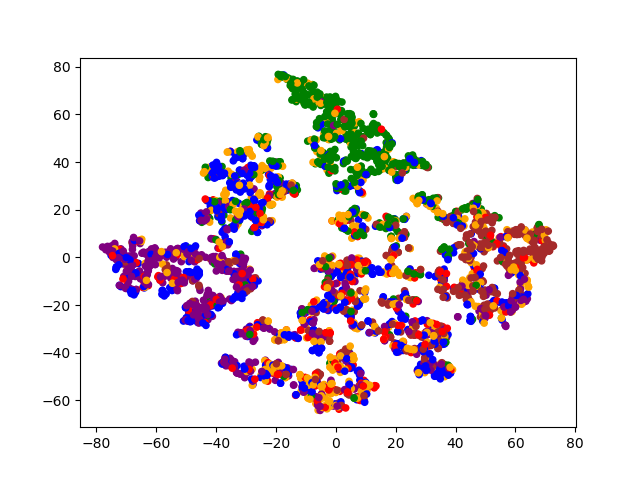}
        \caption{VGAE}
    \end{subfigure}
    \caption{CiteSeer Embeddings Visualization}
    \label{fig:citeseer_vis}
\end{figure}

\begin{figure}[t!]
    \centering
    \begin{subfigure}[b]{0.325\textwidth}
        \centering
        \includegraphics[width=\textwidth]{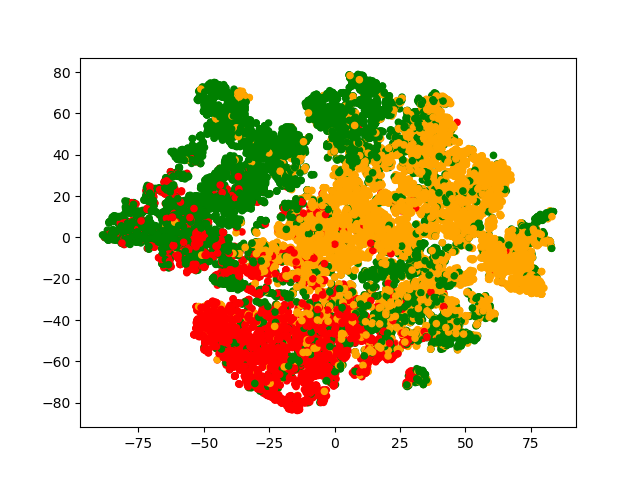}
    \caption{GCN}
    \end{subfigure}
    ~
    \begin{subfigure}[b]{0.325\textwidth}
        \centering
        \includegraphics[width=\textwidth]{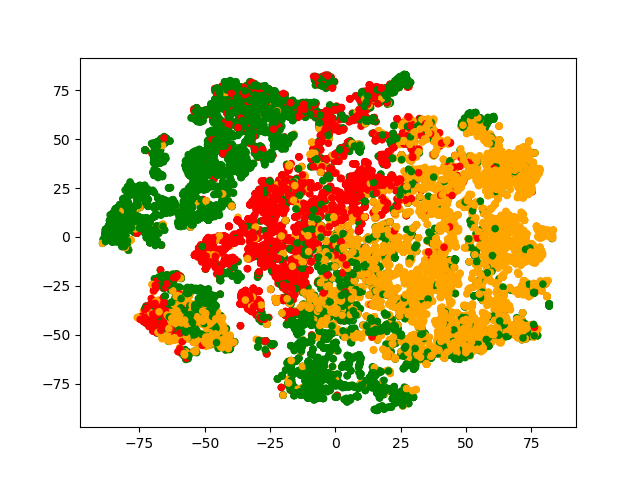}
        \caption{GraphSage}
    \end{subfigure}
    ~
    \begin{subfigure}[b]{0.325\textwidth}
        \centering
        \includegraphics[width=\textwidth]{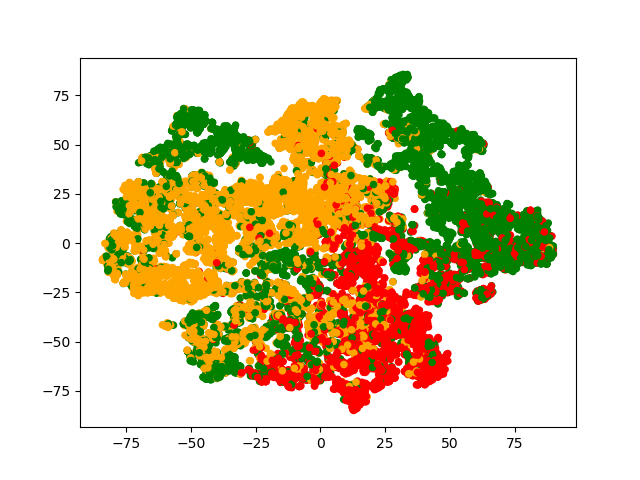}
        \caption{GAT}
    \end{subfigure}
    ~
    \begin{subfigure}[b]{0.325\textwidth}
        \centering
        \includegraphics[width=\textwidth]{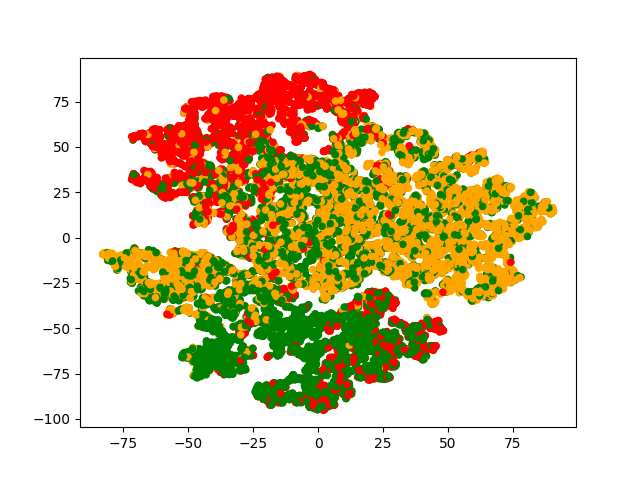}
        \caption{VGAE}
    \end{subfigure}
    \caption{PubMed Embeddings Visualization}
    \label{fig:pubmed_vis}
\end{figure}

\begin{figure}[t!]
    \centering
    \begin{subfigure}[b]{0.325\textwidth}
        \centering
        \includegraphics[width=\textwidth]{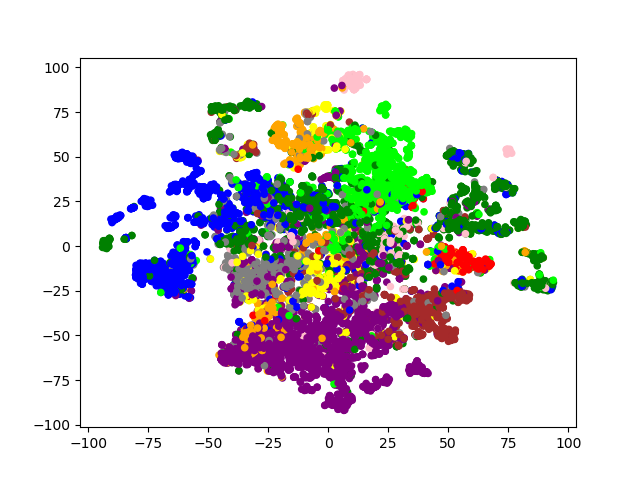}
    \caption{GCN}
    \end{subfigure}
    ~
    \begin{subfigure}[b]{0.325\textwidth}
        \centering
        \includegraphics[width=\textwidth]{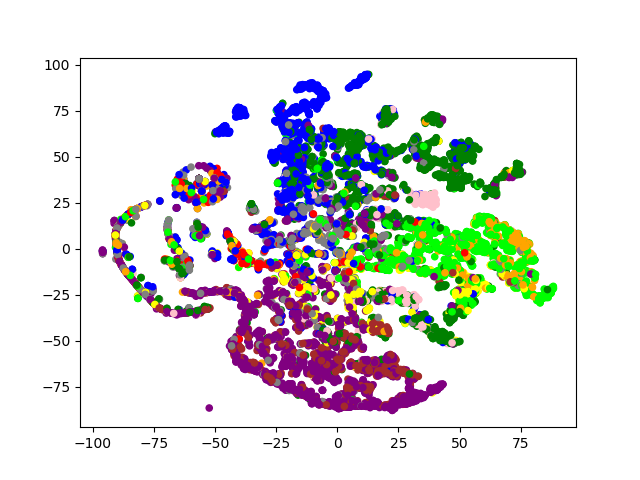}
        \caption{GraphSage}
    \end{subfigure}
    ~
    \begin{subfigure}[b]{0.325\textwidth}
        \centering
        \includegraphics[width=\textwidth]{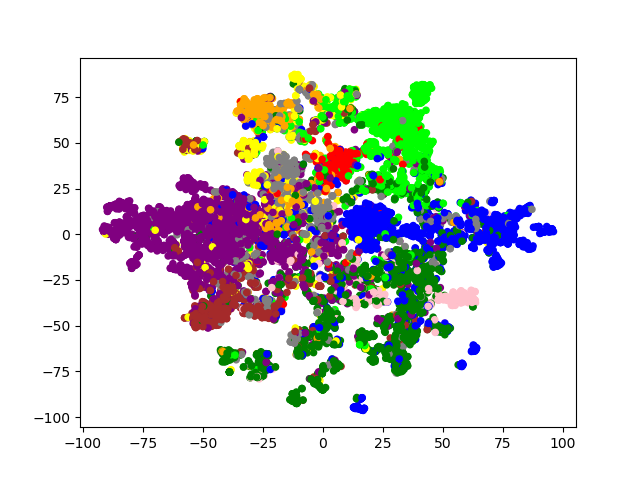}
        \caption{GAT}
    \end{subfigure}
    ~
    \begin{subfigure}[b]{0.325\textwidth}
        \centering
        \includegraphics[width=\textwidth]{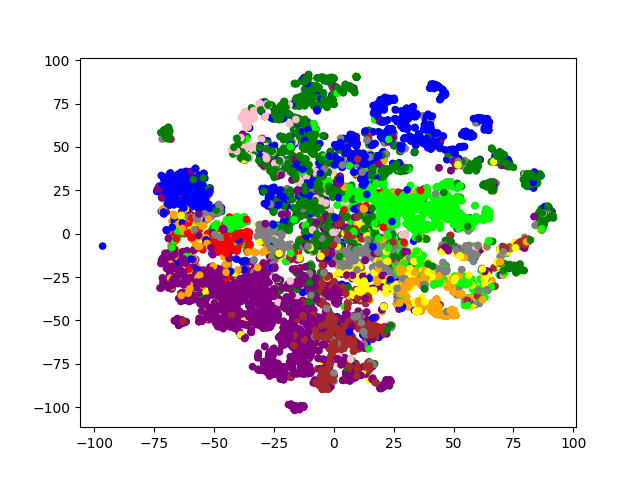}
        \caption{VGAE}
    \end{subfigure}
    \caption{WikiCS Embeddings Visualization}
    \label{fig:wikics_vis}
\end{figure}


\end{document}